\documentclass{article}

 \PassOptionsToPackage{numbers, sort,compress}{natbib}

\usepackage[preprint]{nips_2018}

\usepackage{natbib}
\usepackage[utf8]{inputenc} 
\usepackage[T1]{fontenc}    
\usepackage{hyperref}       
\usepackage{url}            
\usepackage{booktabs}       
\usepackage{amsfonts}       
\usepackage{nicefrac}       
\usepackage{microtype}      

\usepackage{mathrsfs}
\usepackage{amsmath,amsfonts,amssymb,bm}

\usepackage{algorithm}
\usepackage[noend]{algpseudocode}

\usepackage[toc,page]{appendix}

\usepackage{graphicx}
\usepackage{subfigure}

\usepackage[subtle]{savetrees}

\title{On the Importance of Attention in Meta-Learning\\for Few-Shot Text Classification}

\author{
  Xiang Jiang
  \\ Imagia Inc.,\\ Montreal, QC, Canada
  \\ Dalhousie University, \\
  Halifax, NS, Canada\\
  \texttt{\small xiang.jiang@dal.ca}
\And Mohammad Havaei \\
Imagia Inc.,\\ Montreal, QC, Canada\\
\texttt{\small mohammad@imagia.com}
\And Gabriel Chartrand  \\
Imagia Inc.,\\ Montreal, QC, Canada \\
\texttt{\small gabriel@imagia.com}
\AND Hassan Chouaib  \\
Imagia Inc.,\\ Montreal, QC, Canada \\
\texttt{\small hassan.chouaib@imagia.com}
\And Thomas Vincent  \\
Imagia Inc.,\\ Montreal, QC, Canada \\
\texttt{\small thomas.vincent@imagia.com}
\And Andrew Jesson  \\
Imagia Inc.,\\ Montreal, QC, Canada \\
\texttt{\small andrew.jesson@imagia.com}
\AND Nicolas Chapados  \\
Imagia Inc.,\\ Montreal, QC, Canada \\
\texttt{\small nic@imagia.com}
\And Stan Matwin
\\
Dalhousie University, \\
Halifax, NS, Canada \\
\texttt{\small stan@cs.dal.ca}
}

\begin{document}

\maketitle

\begin{abstract}
Current deep learning based text classification methods are limited by their ability to achieve fast learning and generalization when the data is scarce.
We address this problem by integrating a meta-learning procedure that uses the knowledge learned across many tasks as an inductive bias towards better natural language understanding.
Based on the Model-Agnostic Meta-Learning framework (MAML), we introduce the Attentive Task-Agnostic Meta-Learning (ATAML) algorithm for text classification.
The essential difference between MAML and ATAML is in the separation of task-agnostic representation learning and task-specific attentive adaptation.
The proposed ATAML is designed to encourage task-agnostic representation learning by way of task-agnostic parameterization and facilitate task-specific adaptation via attention mechanisms.
We provide evidence to show that the attention mechanism in ATAML has a synergistic effect on learning performance.
In comparisons with models trained from random initialization, pretrained models and meta trained MAML, our proposed ATAML method generalizes better on single-label and multi-label classification tasks in miniRCV1 and miniReuters-21578 datasets.
\end{abstract}

\section{Introduction}
\label{intro}
Deep neural networks have shown great success in learning representations from data,
but effective training of a deep neural network requires a large number of training examples and many gradient-based optimization steps.
This is mainly owing to a lack of prior knowledge when solving a new task.
Meta-learning or ``learning to learn'' \cite{schmidhuber:1987:srl,mitchell1993explanation,vilalta2002perspective} addresses this limitation by acquiring meta-knowledge from the learning experience across many tasks.
The knowledge acquired by the meta-learner provides inductive bias \cite{thrun1998lifelong} that gives rise to sample-efficient fast learning algorithms.

Previous work on deep learning based meta-learning can be summarized into four categories: learning representations that encourage fast adaptation on new tasks \cite{finn2017model,finn2017one}, learning universal learning procedure approximators by supplying training examples to the meta-learner that outputs predictions on the testing examples \cite{hochreiter2001learning,vinyals2016matching,santoro2016meta,mishra2017meta}, learning to generate model parameters conditioned on training examples \cite{gomez2005evolving,munkhdalai2017meta,ha2016hypernetworks}, and learning optimization algorithms to exploit structures in related problems \cite{bengio1992optimization, ravi2016optimization,andrychowicz2016learning,li2017learning}.

Although considerable research has been devoted to meta-learning, research until now has tended to focus on image classification and reinforcement learning, while less attention has been paid to text classification.
In this work, we propose a meta-learning algorithm notably designed for text classification.
The proposed method is based on Model-Agnostic Meta-Learning 
\citep[MAML; see][]{finn2017model} that explicitly guides optimization towards adaptive representations.
While MAML does not discriminate different levels of representations and adapts all parameters for a new task, we introduce Attentive Task-Agnostic Meta-Learner (ATAML) that learns task-agnostic representation while fast-adapting attention parameters to distinguish different tasks.

In effect, ATAML involves two levels of learning: a meta-learner that learns across many tasks to obtain task-agnostic representation in the form of a convolutional or recurrent network, and a base-learner that optimizes the attention parameters of each task for fast adaptation.
Crucially, ATAML takes into account of the importance of attention in document classification and aims to encourage task-specific attentive adaptation while learning task-agnostic text representations.

We introduce a smaller version of the RCV1 and Reuters-21578 dataset---miniRCV1 and miniReuters-21578---tailored to few-shot text classification, and we show that on these datasets, our method leads to significant improvements when compared with randomly initialized, pretrained and MAML-learned models.
We also analyze the impact of architectural choices for representation learning and show the effectiveness of dilated convolutional networks for few-shot text classification.
Furthermore, the findings in both datasets support our claim on the importance of attention in text-based meta-learning.

The contribution of this work is threefold:
\begin{itemize}
\item We propose a new meta-learning algorithm ATAML for text classification that separates task-agnostic representation learning and task-specific attentive adaptation.
%
%
%
%

\item We show that attentive base-learner together with task-agnostic meta-learner generalizes better.
\item We provide evidence as to how attention helps representation learning in ATAML.
\end{itemize}

\section{Model-Agnostic Meta-Learning}
In this section, we follow the same meta-learning problem formulation \cite{ravi2016optimization}
and revisit the MAML \cite{finn2017model} algorithm which we adapt for text classification in Section \ref{sec:method}.

Model-Agnostic Meta-Learning \citep{finn2017model} is a meta-learning algorithm that aims to learn representations that encourage fast adaptation across different tasks.
The meta-learner and base-learner share the same network structure, and the parameters learned by the meta-learner are used to initialize the base-learner on any given task.

To form an ``episode'' \citep{vinyals2016matching} to optimize the meta-learner, we first sample a set of tasks $\{\mathcal{D}_{1},\mathcal{D}_{2},\dots,\mathcal{D}_{S}\}$ from the meta-training set $\mathscr{D}_{\mathrm{meta-train}}$, where $\mathcal{D}_{i}=\{ \mathcal{D}_i^{\mathrm{train}} , \mathcal{D}_i^{\mathrm{test}} \}$.
For a meta-learner parameterized by  $\theta$, we compute its adapted parameters $\theta_{i}$ for each sampled task $\mathcal{D}_{i}$:
\begin{equation}
\theta_{i} \leftarrow \theta - \beta_{\mathrm{T}} \nabla_{\theta}\mathcal{L}(\mathcal{D}_{i}^{\mathrm{train}};\theta),
\end{equation}
where $\beta_{\mathrm{T}}$ is the step size of the gradient.
%
%
%
%
The adapted parameters $\theta_{i}$ are task-specific and tell us the effectiveness of $\theta$ as to whether it can achieve generalization through one or a few additional gradient steps.
The meta-learner's objective is hence to minimize the generalization error of $\theta$ across all tasks:
\begin{equation}
\theta^{\ast}=\mathrm{argmin}_{\theta}\sum_{\mathcal{D}_i \sim \mathscr{D}_{\mathrm{meta-train}}}\mathcal{L}(\mathcal{D}_i^{\mathrm{test}}; \theta_{i}).
\end{equation}
Note that the meta-learner is not aimed at explicitly optimizing the task-specific parameters $\theta_{i}$.
Rather, the objective of the meta-learner is to optimize the representation $\theta$ so that it can lead to good task-specific adaptations $\theta_{i}$ with a few gradient steps.
In other words, the goal of fast learning is integrated into the meta-learner's objective.

The meta-learner is optimized by backpropagating the error through the task-specific parameters $\theta_i$ to their common pre-update parameters $\theta$.
The gradient-based updating rule is:
\begin{equation}
\theta \leftarrow \theta - \beta_{\mathrm{M}}  \nabla_{\theta}\sum_{\mathcal{D}_i \sim \mathscr{D}_{\mathrm{meta-train}}}\mathcal{L}(\mathcal{D}_i^{\mathrm{test}}; \theta_{i}),
\end{equation}
where $\beta_{\mathrm{M}} $ is the learning rate of the meta-learner.
The meta-learner performs slow learning at the meta-level across many tasks to support fast learning on new tasks.
At meta-test time, we initialize the base-learner's parameters from the meta-learned representation $\theta^{\ast}$ and fine-tune the base-learner using gradient descent on task $\mathcal{D}_i^{\mathrm{train}} \sim \mathscr{D}_{\mathrm{meta-test}}$.
The meta learner is evaluated on $\mathcal{D}_i^{\mathrm{test}} \sim \mathscr{D}_{\mathrm{meta-test}}$.

MAML works with any differentiable neural network structure and has been applied to various tasks including regression, image classification, reinforcement learning and imitation learning.
Extensions of MAML include learning the base-learner's learning rate \cite{li2017meta} and applying a bias transformation
to concatenate a vector of parameters to the hidden layer of the base-learner \cite{finn2017one}.
It is also theorized that MAML has the same expressive power as other universal learning procedure approximators and generalizes well to out-of-distribution tasks \cite{finn2017meta}.

\section{Few-Shot Text Classification}
%
%
Few-shot learning is commonly characterized as $N$-way $K$-shot, or $N$-class $K$-shot learning, which contains $N$ classes with only $K$ examples for each class.
Few-shot learning is often accomplished by making use of the knowledge learned from a collection of tasks at an earlier time, and it has made rapid progress on image classification problems \cite{fei2006one,lake2015human,koch2015siamese}, often represented by the Omniglot \cite{lake2011one} and MiniImageNet \cite{vinyals2016matching,ravi2016optimization} datasets.

We extend few-shot learning from image classification to the text classification domain, with the goal to learn a text classification model from a few examples.
Many important problems require learning text classification models from small amounts of data.
As an example, predicting if a person is likely to commit suicide could save many lives, but it is often difficult to collect psychiatric case histories and suicide notes \cite{shneidman1956clues,matykiewicz2012effect}.
Furthermore, in the biomedical domain, active learning can be used to classify clinical text data so as to reduce the burden of annotation \cite{figueroa2012active}. The ability to achieve fast and effective learning from a few annotated examples can jump-start the active learning process and improve the convergence of active learning, thereby maximizing the efficiency of human involvement.


A great body of research in natural language processing emphasizes on the importance of attention in a variety of tasks \cite{shen2018bi,lin2017structured, vaswani2017attention}. These papers show that attention is able to retrieve task-specific representation across a sequence of text encodings from CNN or LSTM to obtain a task specific representation of the input.
Attention could help decompose the contents of a document into ``subproblems'' \cite{parikh2016decomposable} thus producing task-specific representations; this ability to decompose text encodings also allows us to learn shared representation across tasks.

In the context of meta-learning for few shot text classification, we empirically show that there is a synergistic relationship between meta-learning a shared text embedding across tasks and task-specific representation through attention. Intuitively, by constraining the text embedding parameters to be shared across different tasks in an episode, attention learns to be more task-specific and better decompose the document according to the task at hand.

\section{Attentive Task-Agnostic Meta-Learning}
\label{sec:method}
\subsection{The Attentive Base Learner}
\label{sec:base_learner}
The base-learner is a neural network trained on each text classification task $\mathcal{D}$ under a loss function $\mathcal{L}$.
The base-learner reads the $T$-word input document $\mathbf{x}=\left [ x_1,x_2,\dots,x_T \right ]$, where $x_t$ denotes the $t$-th word,
\begin{equation}
\label{equ:bilstm}
\mathbf{s}_t=f(x_t; \theta_\mathrm{E}).
\end{equation}
The base learner in~\eqref{equ:bilstm}
encodes the input sequence $\mathbf{x}$ to a corresponding sequence of states $\left [ \mathbf{s}_1,\mathbf{s}_2,\dots,\mathbf{s}_T \right ]$, where $f$ can take the form of a recurrent or convolutional network with parameters $\theta_\mathrm{E}$.

We then apply content-based attention mechanism \cite{bahdanau2014neural,hermann2015teaching,graves2014neural,sukhbaatar2015end} that enables the model to focus on different aspects of the document.
The specific attention formulation used here is defined in~\eqref{equ:attention_layer} and belongs to a type of feedforward attention \cite{raffel2015feed},
\begin{equation}
\label{equ:attention_layer}
\alpha_t=\bm{\theta}_{\mathrm{ATT}}^\intercal \mathbf{s}_t
,\qquad 
\mathbf{s}'_t = \alpha_t \mathbf{s}_t 
,\qquad 
\mathbf{c}=\frac{1}{T}\sum_{t=1}^{T}\mathbf{s}'_t,
\end{equation}
where $\bm{\theta}_{\mathrm{ATT}}$ represents the attention parameter vector.
For each memory state $\mathbf{s}_t$, we calculate its inner product with the attention parameter, resulting in a scalar $\alpha_t$.
The scalar $\alpha_t$ rescales each state $\mathbf{s}_t$ into $\mathbf{s'}_t$, and these are averaged to obtain the final representation $\mathbf{c}$ of a document.
The attention retrieves relevant information from a document and offers interpretability into the model behavior by explaining the importance of each word, through attention weight $\alpha_t$, that contributes to the final prediction.

Once an input document $\mathbf{x}$ is encoded into the vectorized representation $\mathbf{c}$, we apply a softmax classifier parameterized by $\bm{\theta}_W$ to obtain the predictions $\hat{y}$.
The softmax classifier is replaced by a set of sigmoid classifiers if the labels are not mutually exclusive in multi-label classification,
\begin{equation}
\hat{y}=\mathrm{softmax}(\mathbf{c}; \bm{\theta}_W) \quad \mathrm{or} \quad \hat{y}=\mathrm{sigmoid}(\mathbf{c}; \bm{\theta}_W).
\end{equation}
 
\subsection{The Attentive Task-Agnostic Meta-Learner}
\label{sec:meta_learner}
ATAML learns to obtain common representations that can be shared across different tasks while having the fast learning ability to quickly adapt to new tasks.
In contrast with MAML, which does not make any distinction between different parameters in the meta-learner, the proposed ATAML splits all parameters $\theta$ into into two disjoint sets, shared parameters  $\theta_\mathrm{E}$ and task-specific parameters $\theta_\mathrm{T}$, and employs discriminative strategies in the meta-training and meta-testing phrases.
The shared parameters $\theta_\mathrm{E}$ are aimed at representation learning while the task-specific parameters $\theta_\mathrm{T}$ are aimed at capturing task-specific information for classification.

\subsubsection{Meta Training}
\begin{algorithm}
\caption{Attentive Task-Agnostic Meta-Learner}\label{alg:maml_text}
\label{alg:MAML}
\begin{algorithmic}[1]
\Require $\mathscr{D}_{\mathrm{meta-train}}$: the meta-train set
\Require $N$-way $K$-shot learning
\Require $S$ classification tasks for each training episode
\Require $\beta_{\mathrm{T}}, \beta_{\mathrm{M}}$: task and meta level learning rate
\Require $\theta_{\mathrm{E}}$: shared parameters for representation learning
\Require $\theta_\mathrm{T} = \{\bm{\theta}_W, \bm{\theta}_{\mathrm{ATT}}    \}$: parameters to be adapted at the task level
\State randomly initialize $\theta_{\mathrm{E}}$ and $\theta_{\mathrm{T}}$ \Comment{Initialize all parameters}
\While{not done}
\State Sample $S$ tasks: $\mathcal{D}_i \sim \mathscr{D}_{\mathrm{meta-train}}$\Comment{Sample tasks for meta-training}
\ForAll {$\mathcal{D}_i$}
\State $\theta_{\mathrm{T},i} = \theta_\mathrm{T} - \beta_{\mathrm{T}} \nabla_{\theta_{\mathrm{T},i}}\mathcal{L}(\mathcal{D}_i^{\mathrm{train}};\theta_{\mathrm{T},i})$ \Comment{Get task-specific parameters}

\EndFor
\State $\mathcal{L}_{\mathrm{meta}} = \sum_{\mathcal{D}_i}\mathcal{L}(\mathcal{D}_i^{\mathrm{test}}; \{ \theta_{\mathrm{T},i}, \theta_{\mathrm{E}} \})$\Comment{Get loss of the meta-learner}
\State $\theta_\mathrm{T} \gets  \theta_\mathrm{T} - \beta_{\mathrm{M}} \nabla_{\theta_\mathrm{T}}\mathcal{L}_{\mathrm{meta}}$\Comment{Update task-specific parameters}
\State $\theta_\mathrm{E} \gets  \theta_\mathrm{E} - \beta_{\mathrm{M}} \nabla_{\theta_\mathrm{E}}\mathcal{L}_{\mathrm{meta}}$\Comment{Update shared parameters}
\EndWhile
\end{algorithmic}
\end{algorithm}

The Attentive Task-Agnostic Meta-Learning training algorithm is described in Algorithm \ref{alg:MAML}.
We use $\theta$ to denote all parameters of the model ($\theta = \{\bm{\theta}_W, \bm{\theta}_{\mathrm{ATT}},  \theta_{\mathrm{E}}  \}$), which is divided into shared parameters $\theta_\mathrm{E}$ and task-specific parameters $\theta_\mathrm{T}$, where $\theta_\mathrm{T} = \{\bm{\theta}_W, \bm{\theta}_{\mathrm{ATT}}    \}$.

To create one meta-training ``episode'' \citep{vinyals2016matching}, we sample $S$ tasks from $\mathscr{D}_{\mathrm{meta-train}}$ and optimize the model towards fast learning across all sampled tasks $\left [\mathcal{D}_1, \mathcal{D}_2, \dots,\mathcal{D}_S  \right ]$.
As we are sampling random tasks from $\mathscr{D}_{\mathrm{meta-train}}$ in each meta-training iteration, the goal of the meta-learner is to obtain task-agnostic representation $\theta_\mathrm{E}$ that is reusable for different tasks.

For every task $\mathcal{D}_i$ in the meta-training iteration, we only update the task-specific parameters in the base-learner that are initialized with $\theta_\mathrm{T}$ and updated to $\theta_{\mathrm{T},i}$ using task-specific gradients $\nabla_{\theta_{\mathrm{T},i}}\mathcal{L}(\mathcal{D}_i^{\mathrm{train}};\theta_{\mathrm{T},i})$.
We further calculate the expected loss according to the post-update parameters that is composed of the task-specific fast weights $\theta_{\mathrm{T},i}$ and shared slow weights $\theta_\mathrm{E}$,
\begin{equation}
\mathcal{L}_{\mathrm{meta}} = \sum_{\mathcal{D}_i}\mathcal{L}(\mathcal{D}_i^{\mathrm{test}}; \{ \theta_{\mathrm{T},i}, \theta_{\mathrm{E}} \}).
\end{equation}
The resulting loss $\mathcal{L}_{\mathrm{meta}}$ can be understood as the loss of the meta-learner.
$\mathcal{L}_{\mathrm{meta}}$ gives us an evaluation measure on how well the task-specific parameters $\theta_{\mathrm{T}}$ can adapt across all the sampled tasks $\mathcal{D}_i$, together with a measure on how well the shared parameters $\theta_{\mathrm{E}}$ can be reused across all tasks.
%
%
The meta-optimization therefore consists of minimizing $\mathcal{L}_{\mathrm{meta}}$ with respect to all parameters $\theta$ towards optimizing the model's adaptability and re-usability across different tasks.
The meta-training iterations are repeated until the model converges, and the resulting parameters $\theta$ are then used as initialization at meta-test time.

\subsubsection{Meta Testing}
Meta testing involves evaluating on the meta-learned model on the meta-test set $\mathscr{D}_{\mathrm{meta-test}}$ by fine-tuning on $\mathcal{D}_i^{\mathrm{train}}$ and test on $\mathcal{D}_i^{\mathrm{test}}$, where $\mathcal{D}_i \sim \mathscr{D}_{\mathrm{meta-test}}$.
We introduce a meta testing approach that freezes the shared representation learning parameters $\theta_{\mathrm{E}}$ and only applies gradient on the task-specific parameters $\theta_{\mathrm{T}}$.
In contrast to fine-tuning all parameters for a new task, this discriminative meta-testing procedure is more coherent with the stratified meta-training strategy.
It also provides regularization of few-shot learning that improves generalization.

\section{Experiments}
We provide three sets of empirical evaluations on the single-label miniRCV1, multilabel miniRCV1 and miniRCV1miniReuters-21578 datasets to analyze the proposed meta-learning framework.

\label{sec:eval_homogeneous}

\subsection{The Base Learners}
We use Temporal Convolutional Networks (TCN), which is a type of dilated convolution \cite{van2016wavenet}, as the base learner.
We have also conducted experiments with bidirectional LSTM \cite{schuster1997bidirectional} as the base learner. Details on those experiments as well as the LSTM architecture are included in the Appendix due to lack of space.

The TCN contains two layers of dilated causal convolutions with filter size 3 and dilation rate 3.
Each convolutional layer is followed by a Leaky Rectified Linear Unit \cite{maas2013rectifier} with negative slope rate 0.01, which is followed by 50\%  dropout \cite{srivastava2014dropout}.
For word representation, we use 300 dimensional Glove embeddings \cite{pennington2014glove}.
For optimization, we use Adam optimizer \cite{kingma2014adam}.
For the loss function, we use categorical cross entropy error when each document contains only one label, and use sigmoid cross entropy error when each document may contain multiple labels.
Although it is common to use threshold calibration algorithms for multilabel classification, we use the constant 0.5 as prediction threshold in order to reduce the impact of external algorithms.

\subsection{Data}
Reuters Corpus Volume I (RCV1) is an archive of news stories for research on text categorization \cite{lewis2004rcv1}.
We create two versions of the miniRCV1 dataset by selecting a subset from the full RCV1 dataset to study the effect of few-shot learning in text classification:
\begin{enumerate}
\item \textit{miniRCV1 for single-label classification} consisting of the 55 second-level topics as target classes.
We sample 20 documents from each class which is further divided into a training set that contains 5 documents and a testing set that contains 15 documents.
Documents with overlapping topics are removed to ensure each document contains a single label.
\item \textit{miniRCV1 for multi-label classification} consisting of 102 out of 103 non-mutually exclusive labels.
Each document is associated with a set of labels and we exclude one label that only appeared once in the corpus.
We sample about 20 documents for each class and divide them into training and testing sets in a similar manner.
It is worthwhile to mention that, due to the inherent properties of multi-labeled data \cite{zhang2014review}, some classes may contain more examples than others classes.
\end{enumerate}
Similar to miniRCV1, we create a smaller version of the Reuters-21578 dataset by selecting about 20 examples for each label.

\subsection{Few-shot Learning Setup}
At the meta-level, we divide all classes into meta-train, meta-validation and meta-test sets. 
In the $N$-way $K$-shot setup, during meta-training, we randomly sample $N$ classes among the meta-training set where each class contains $K$ training examples.
At meta-test time, we randomly sample $N$ classes among the meta-test set and calculate evaluation statistics across many runs.
We evaluate 5-way 1-shot, 5-way 5-shot, 10-way 1-shot and 10-way 5-shot learning for both single-label and multi-label classification.
The single-label classification task is evaluated on classification accuracy; the multi-label classification task is evaluated on micro and macro F1-scores, which are intended to measure the average F1-scores across all labels.
They differ in that, micro-average gives equal weights to each example regardless of label imbalance, whereas macro-average treats different labels equally.

\subsection{Results and Discussion}

As with other meta-learning paradigms we consider two baselines:~models trained from random initialization, i.e., ``random'', and~models pretrained across many sampled meta-train tasks, i.e., ``pretrained''. In addition, we also compare our proposed ATAML framework with MAML under similar architecture. Our experiments show that while MAML achieves better accuracies compared to the aforementioned baselines, ATMAL significantly outperforms MAML in all 1-shot learning experiments. Table~\ref{tab:rcv1_acc}, Table~\ref{tab:rcv1_multilabel} and Table~\ref{tab:r21578} summarize these results on single-label miniRCV1, multi-label miniRCV1 and multi-label miniReuters-21578 experiments, wherein ``Meta'' denotes the type of meta learner, ``Base'' denotes the type of base learner,
``(A)'' denotes models trained with attention and the bold numbers highlight the best performing ones at 95\% confidence interval.

\paragraph{The difficulty of few-shot learning.}
Few-shot text classification is a challenging task as text data contain rich information from various aspects which are difficult to ascertain from a few training examples.
This difficulty is manifested in the poor testing performances when trained from random initialization.
Meanwhile, in both multi-label classification tasks, the TCN models perform much better when we increase the number of training examples from 1 to 5 examples per class.
Furthermore, we show in the Appendix that, classic machine learning algorithms, such as support vector machine, naive Bayes multinomial and K-nearest neighbors, as well as document embedding algorithms, such as doc2vec~\cite{levine1985effect} and doc2vecC~\cite{chen2017efficient}, also suffer from data scarcity in few-shot learning.

\paragraph{The difficulty of pretraining in few-shot learning.}
We empirically find it generally ineffective to make use of pretrained models in few-shot learning.
This can be explained by the ``contradictory outputs'' of the pretraining tasks \cite{finn2017model}.
Put differently, as each task contains a small number of examples, when we pretrain the model from many tasks in the meta-training set, the sampled tasks provide contradictory supervisory signals to the classifier, hence making it difficult to pretrain effectively.

\paragraph{Why does pretrained $10$-way $K$-shot TCN models perform so poorly?}
In multi-label classification tasks, some labels appear less frequently in the training data . This label imbalance causes uncalibrated output probabilities when using the constant 0.5 as prediction threshold.
Some pretrained models performs worse than random guesses because its output probabilities are not well distributed.

\paragraph{The effect of meta learning.}
From all three experiments, the empirical results demonstrate the basic MAML with attentive base learners performs notably better than the non-meta-learned baselines.
More importantly, the proposed ATAML algorithm offers further improvements that are statistically significant in all the 1-shot learning experiments.
These empirical findings support the need for meta-learning in few-shot text classification.
That being the case, the empirical findings further support the importance of learning task-agnostic representations together with task-specific attentive adaptations.

\begin{table}[t]
\renewcommand{\arraystretch}{1.15}
\centering
\caption{Comparing single-label classification accuracies between baselines and ATAML on miniRCV1}\smallskip
\vskip -0.15in
\label{tab:rcv1_acc}
\begin{center}
\begin{small}
\begin{tabular}{llcccc}
\toprule
\multicolumn{2}{c}{Method }                    & \multicolumn{2}{c}{5-way Accuracy} & \multicolumn{2}{c}{10-way Accuracy} \\ \cmidrule(r{4pt}){1-2} \cmidrule(l){3-4} \cmidrule(l){5-6}
      Meta & Base                           & 1-shot     & 5-shot     & 1-shot     & 5-shot    \\ \midrule
random & TCN (A) & 41.52\% & 65.64\% & 28.32\% & 45.12\% \\ 
pretrained & TCN (A) & 24.06\% & 57.08\% & 18.60\% & 45.85\% \\ 
MAML & TCN (A) & 47.09\% & \textbf{72.65\%} & 31.57\% &\textbf{ 62.75\%} \\
\cmidrule(l){1-6}
ATAML & TCN (A) & \textbf{54.05\%} & \textbf{72.79\%} & \textbf{39.48\%} & \textbf{61.74\%} \\ 
\bottomrule
\end{tabular}
\end{small}
\end{center}
\vskip -0.15in
\end{table}

\begin{table}[t]
\renewcommand{\arraystretch}{1.15}
\centering
\caption{Comparing multi-label classification outcomes between baselines and ATAML on miniRCV1}\smallskip
\vskip -0.12in
\centering
\label{tab:rcv1_multilabel}
\begin{small}
\begin{tabular}{llcccccccc}
\toprule
\multicolumn{2}{c}{Method }                   & \multicolumn{2}{c}{5-way Micro-F1} & \multicolumn{2}{c}{10-way Micro-F1}  & \multicolumn{2}{c}{5-way Macro-F1} & \multicolumn{2}{c}{10-way Macro-F1}
\\\cmidrule(r{4pt}){1-2} \cmidrule(r{4pt}){3-4} \cmidrule(l){5-6} \cmidrule(r{4pt}){7-8} \cmidrule(l){9-10}
             Meta & Base           & 1-shot     & 5-shot     & 1-shot     & 5-shot  & 1-shot     & 5-shot     & 1-shot     & 5-shot    \\ \midrule
random & TCN (A) & 38.9\% & 60.9\% & 40.6\% & 45.6\% & 31.4\% & 55.7\% & 22.9\% & 33.1\% \\ 
pretrained & TCN (A) & 26.9\% & 55.8\% & 33.5\% & 52.1\% & 17.0\% & 51.5\% & 14.9\% & 41.4\% \\ 
MAML & TCN (A) & 52.3\% & \textbf{69.1\%} & 44.9\% & 58.6\% & 43.2\% & \textbf{64.3\%} & 27.7\% & \textbf{48.4\%} \\ 
\cmidrule(l){1-10}
ATAML & TCN (A) & \textbf{59.7\%} & \textbf{71.1\%} & \textbf{50.7\%} & \textbf{61.3\%} & \textbf{54.3\%} & \textbf{65.0\%} & \textbf{38.5\% }& \textbf{49.2\%} \\ 

\bottomrule
\end{tabular}
\end{small}
\vskip -0.12in
\end{table}

\begin{table}[t]
\renewcommand{\arraystretch}{1.15}
\centering
\caption{Comparing multi-label classification between baselines and ATAML on miniReuters-21578}\smallskip
\vskip -0.12in
\centering
\label{tab:r21578}
\begin{small}
\begin{tabular}{llcccccccc}
\toprule
\multicolumn{2}{c}{Method }                   & \multicolumn{2}{c}{5-way Micro-F1} & \multicolumn{2}{c}{10-way Micro-F1}  & \multicolumn{2}{c}{5-way Macro-F1} & \multicolumn{2}{c}{10-way Macro-F1}
\\\cmidrule(r{4pt}){1-2} \cmidrule(r{4pt}){3-4} \cmidrule(l){5-6} \cmidrule(r{4pt}){7-8} \cmidrule(l){9-10}
             Meta & Base           & 1-shot     & 5-shot     & 1-shot     & 5-shot  & 1-shot     & 5-shot     & 1-shot     & 5-shot    \\ \midrule
random & TCN (A) & 38.2\% & 66.0\% & 25.1\% & 44.9\% & 30.6\% & 55.0\% & 17.9\% & 33.6\%  \\ 
pretrained & TCN (A) & 23.5\% & 50.3\% & 18.4\% & 49.1\% & 16.4\% & 37.8\% & 12.0\% & 37.3\%  \\ 
MAML & TCN (A) &52.4\% & \textbf{74.1\%} & 38.1\% & \textbf{61.2\%} & 44.3\% & 64.3\% & 29.9\% & \textbf{51.2\%} \\
\cmidrule(l){1-10}
ATAML & TCN (A) & \textbf{66.3\%} & \textbf{76.5\%} & \textbf{42.6\% }& \textbf{60.8\%} & \textbf{60.9\%} & \textbf{69.4\%} & \textbf{34.9\%} & \textbf{51.2\%}  \\ 
\bottomrule
\end{tabular}
\end{small}
\vskip -0.12in
\end{table}


\subsection{The Importance of Attention}
\label{sec:importance_of_attention}
The importance of attention lies in its synergistic effect on the meta learners.
Under the same meta learning framework, introducing attention to the base learner leads to improved generalization when compared with non-attentive base learners.
From empirical results shown in Table~\ref{tab:ablation}, we find MAML trained with attention performs better than MAML without attention.
We have similar findings on the mini-RCV1 experiments detailed in the Appendix.

Furthermore, the proposed ATAML framework implicitly evaluates the learned representations by freezing the meta-learned representation parameters when fine-tuning a new task.
Under those circumstances, well trained representation will facilitate the learning of a new task while poorly trained representation will prohibit effective adaptations.
From empirical studies, ATAML performs the best across all experiments when we use TCN as the base learner.


If only focusing on base learners that are equipped with attention mechanisms, we find that although MAML provides reasonable improvements from the baselines, models trained with ATAML offer substantial improvements in generalization when compared with the rest of the models.
This hints at the benefits brought by shared representation learning and discriminative fine-tuning.

Putting all these together, we empirically find that both the attention mechanism and the meta learner are crucial components for good generalization in few-shot text classification.
To better understand the representation learning procedure as well as the role of attention in meta training, we undertake ablation studies to provide further insights into ATAML.

\begin{table}[t]
\renewcommand{\arraystretch}{1.15}
\centering
\caption{Ablation studies on miniReuters-21578 for multi-label classification}\smallskip
\centering
\label{tab:ablation}
\begin{small}
\begin{tabular}{llcccccccc}
\toprule
\multicolumn{2}{c}{Method }                   & \multicolumn{2}{c}{5-way Micro-F1} & \multicolumn{2}{c}{10-way Micro-F1}  & \multicolumn{2}{c}{5-way Macro-F1} & \multicolumn{2}{c}{10-way Macro-F1}
\\\cmidrule(r{4pt}){1-2} \cmidrule(r{4pt}){3-4} \cmidrule(l){5-6} \cmidrule(r{4pt}){7-8} \cmidrule(l){9-10}
             Meta & Base           & 1-shot     & 5-shot     & 1-shot     & 5-shot  & 1-shot     & 5-shot     & 1-shot     & 5-shot    \\ \midrule
             random & E (A) & 36.7\% & 66.1\% & 25.2\% & 49.1\%  & 29.2\% & 55.0\% & 18.2\% & 36.8\% \\
             MAML & E (A) & 44.9\% & 72.3\% & 26.4\% & 59.2\% & 35.6\% & 61.7\% & 19.6\% & 47.4\% \\
             \cmidrule(l){1-10}
             MAML & TCN & 26.4\% & 65.7\% & 11.4\% & 44.5\% & 19.1\% & 52.7\% & 7.6\% & 31.2\%  \\ 

MAML & TCN (A) &52.4\% & \textbf{74.1\%} & 38.1\% & \textbf{61.2\% } & 44.3\% & \textbf{64.3\%} & 29.9\% & \textbf{51.2\%} \\

\cmidrule(l){1-10}
ATAML & TCN (A) & \textbf{66.3\%} & \textbf{76.5\%} & 42.6\% & \textbf{60.8\% }& \textbf{60.9\%} & \textbf{69.4\%} & 34.9\% & \textbf{51.2\% }\\ 
ATAML & TCN (-) & 62.7\% & \textbf{77.5\%} &\textbf{ 49.5\%} & \textbf{63.7\%} & \textbf{58.3\%} & \textbf{71.1\%} & \textbf{41.6\%} & \textbf{54.2\%} \\

\bottomrule
\end{tabular}
\end{small}

\end{table}
\subsection{Ablation Studies}

With ablation studies we can offer evidence into the need to learn text in a structured manner as opposed to making classifications at the word level alone.
We use ``E (A)'' to denote a base learner where an attention model is directly applied to the word embeddings.
The goal of this model is to extract individual words to make predictions.
This model provides a measure on classification performance if we only take into account individual word-level representations.
The empirical results in Table~\ref{tab:ablation} suggest classifying from word embeddings is inferior to the proposed ATAML model, indicating the need to learn text structures, such as phrase or sentence level representations.
Moreover, learning from only a few examples exacerbates the effect of over-fitting as it is more likely to have spurious correlations at the word level compared with phrase or sentence level.
It is therefore desirable to have the ability to learn text structures.

To analyze the role of attention in meta training, we construct an attention-based meta training strategy where the attention parameters are not updated in each meta training iteration.
Although the attention parameters are not being updated in meta training, they take task-specific fast weights as regular ATAML and these fast weights have direct influence over the gradients of the TCN layers.
The goal of this model is to exploit the fast weights of the attention parameters and examine if this could produce well trained representation without learning attention parameters in meta learning.
This model, denoted as ``TCN(-)'', has similar performance with the regular ATAML models in Table~\ref{tab:ablation}.
Thus, the role of attention in meta training is to facilitate the learning of shared representations, rather than learning attention parameter itself.
In addition, we show in the Appendix that, the proposed ATAMA works better than document embedding approaches that further confirms its ability to aggregate information from substructures.

\subsection{Visualizing Learned Attentions}

\begin{figure}
  \centering
  \includegraphics[scale=0.68]{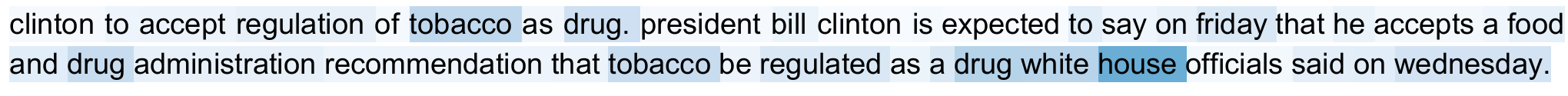}
  \caption{Visualizing attentions learned by MAML TCN(A).}
    \label{fig:attention_MAML}
\end{figure}
\begin{figure}
  \centering
  \includegraphics[scale=0.68]{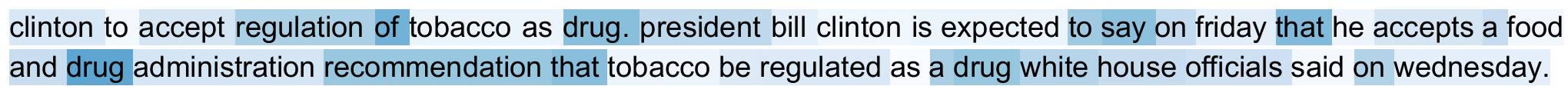}
  \caption{Visualizing attentions learned by ATAML TCN(A).}
    \label{fig:attention_ATAML}
\end{figure}

Figure~\ref{fig:attention_MAML} and Figure~\ref{fig:attention_ATAML} are visualizations of attention obtained by MAML and ATAML, respectively.
The density of the blue color indicates the weights, or importance, for the words to model predictions. 
The target label for this document is ``REGULATION/POLICY'' and both models make correct predictions for this training example. Additional visualization is provided in the Appendix. 

The MAML model illustrated in Figure~\ref{fig:attention_MAML} is over-fitting on the training data and only searches for repetitive words, such as ``tobacco'' and ``drug'', that are merely spurious correlations.
On the other hand, the proposed ATAML suffers less from over-fitting and searches for relevant phrases, such as ``accept regulation of'' and ``recommendation that'', which are relevant to ``REGULATION/POLICY'' for prediction.
This suggests the proposed ATAML is able to discover local text substructures via attention from shared representation learning which has a regularization effect when adapting to new tasks.

\subsection{The Impact of Base Learner}
Current research on meta learning typically use LSTM as a meta-learner, while we experimented with both LSTM and TCN as the base learner.
Although meta learning works with both LSTM and TCN and they all provide improvements from randomly initialized and pretrained models, it is worthwhile to highlight their different properties.
Overall, TCN has faster training speed and generalization when compared with LSTM.
One main problem when using LSTM as the base learner is that, in meta-training, the LSTM saturates at a very early stage owing to difficulties in optimization, and prevents the meta-learner from obtaining sharable representations across different tasks.
The detailed quantitative comparisons are included in the Appendix.

\section{Conclusion}
\label{conclusions}
We propose a meta learning approach that enables the development of text classification models from only a few training examples.
The proposed Attentive Task-Agnostic Meta-Learner encourages the learning of shared representation across different tasks.
The use of attention mechanism is capable of decomposing some text into substructures for task-specific adaptation.
We also found attention facilitates learning text representations that can be shared across different tasks.
The importance of attention in meta-learning for few-shot text classification is clearly supported by our empirical studies on the miniRCV1 and miniReuters-21578 datasets.
We also provided ablation analysis and visualization to get insights into how different components of the model work together.
To the best of our knowledge, this is the first work to raise the question of few-shot text classification. 
Further work should further characterize what  makes a good few-shot text classification algorithm.

\medskip
\small
\bibliography{meta_learning.bib}

\begin{appendices}
\section{Training Details}
We choose two commonly used text classification models as base learners for empirical analysis.
The first base learner is a bidirectional LSTM \cite{schuster1997bidirectional} that contains 128 hidden nodes in each direction.
The second base learner is a dilated convolutional network \cite{van2016wavenet} that contains two layers of dilated causal convolutions \cite{bai2018empirical} with filter size 3 and dilation rate of 3.
Each of the convolutional layer is followed by leaky rectified linear units \cite{maas2013rectifier} and 50\% dropconnect \cite{wan2013regularization}.
We use residual blocks as described in \cite{bai2018empirical}.
For word representation, we use Glove embeddings \cite{pennington2014glove}.
For optimization, we use Adam optimizer \cite{kingma2014adam} and clip the gradients to a maximum L2-norm of 1.0.
For the loss function, we use categorical cross entropy error when each document contains only one label, and use sigmoid cross entropy error when each document may contain multiple labels.
When Temporal Convolutional Network (TCN) is used as the base learner, we experimented with a three-layer architecture but it did not work as good as a two-layer model.
We have also tried batch normalization but they do not provide performance improvements.
We find taking more than one fast gradient steps in meta training improves learning and we use 5 gradient steps in our experiments.

\section{Additional Empirical Results}
\subsection{The Importance of Attention}
In this section, we include additional empirical results for single-label and multi-label miniRCV1 experiments in Table~\ref{tab:rcv1_acc} and Table~\ref{tab:rcv1_multilabel} to show the importance of attention, wherein ``meta'' denotes the type of meta learner, ``Base'' denotes the type of base learner,
``random'' denotes models trained from random initialization, ``pretrained'' denotes models trained from a pretrained model on the meta-training set, 
``(A)'' denotes models trained with attention and the bold numbers highlight the best performing ones at 95\% confidence interval.

The empirical results suggest that attention provides performance improvements regardless of what meta learner or base learner is used.
Given the same meta learning algorithm, adding attention to the base learner always improves model performance.

\begin{table}[t]
\renewcommand{\arraystretch}{1.15}
\centering
\caption{miniRCV1 single-label classification accuracies}\smallskip
\label{tab:rcv1_acc}
\begin{center}
\begin{small}
\begin{tabular}{llcccc}
\toprule
\multicolumn{2}{c}{Method }                    & \multicolumn{2}{c}{5-way Accuracy} & \multicolumn{2}{c}{10-way Accuracy} \\ \cmidrule(r{4pt}){1-2} \cmidrule(l){3-4} \cmidrule(l){5-6}
      Meta & Base                           & 1-shot     & 5-shot     & 1-shot     & 5-shot    \\ \midrule
random &TCN & 26.70\% & 55.43\% & 17.64\% & 41.81\% \\ 
random &TCN (A) & 41.52\% & 65.64\% & 28.32\% & 45.12\%  \\ 
pretrained &TCN & 22.38\% & 37.17\% & 10.67\% & 27.76\% \\ 
pretrained &TCN (A) & 24.06\% & 57.08\% & 18.60\% & 45.85\% \\ 
MAML &TCN & 33.86\% & 61.44\% & 22.55\% & 41.94\% \\ 
MAML & TCN (A) & 47.09\% & \textbf{72.65\%} & 31.57\% & \textbf{62.75\%} \\
ATAML &TCN (A) & \textbf{54.05\%} & \textbf{72.79\%} & \textbf{39.48\%} & \textbf{61.74\%} \\ 
\bottomrule
\end{tabular}
\end{small}
\end{center}
\vskip -0.1in
\end{table}

\begin{table}[t]
\renewcommand{\arraystretch}{1.15}
\centering
\caption{miniRCV1 multi-label classification}\smallskip
\centering
\label{tab:rcv1_multilabel}
\begin{small}
\begin{tabular}{llcccccccc}
\toprule
\multicolumn{2}{c}{Method }                   & \multicolumn{2}{c}{5-way Micro-F1} & \multicolumn{2}{c}{10-way Micro-F1}  & \multicolumn{2}{c}{5-way Macro-F1} & \multicolumn{2}{c}{10-way Macro-F1}
\\\cmidrule(r{4pt}){1-2} \cmidrule(r{4pt}){3-4} \cmidrule(l){5-6} \cmidrule(r{4pt}){7-8} \cmidrule(l){9-10}
             Meta & Base           & 1-shot     & 5-shot     & 1-shot     & 5-shot  & 1-shot     & 5-shot     & 1-shot     & 5-shot    \\ \midrule
random & TCN & 18.7\% & 40.6\% & 30.2\% & 40.9\% & 11.3\% & 36.4\% & 9.9\% & 23.6\%  \\ 
random & TCN (A) & 38.9\% & 60.9\% & 40.6\% & 45.6\% & 31.4\% & 55.7\% & 22.8\% & 33.1\% \\ 
pretrained & TCN & 25.1\% & 36.2\% & 28.2\% & 35.2\% & 17.0\% & 30.1\% & 9.1\% & 20.7\% \\ 
pretrained & TCN (A) & 26.9\% & 55.8\% & 33.5\% & 52.1\% & 17.0\% & 51.5\% & 14.9\% & 41.4\%  \\ 
MAML & TCN & 35.7\% & 45.6\% & 20.5\% & 40.2\% & 22.9\% & 41.9\% & 7.6\% & 27.7\% \\ 
MAML & TCN (A) & 52.3\% & \textbf{69.1\%} & 44.9\% & \textbf{58.6\%} & 43.2\% & 64.3\% & 27.7\% & \textbf{48.4\%} \\ 
SMAML & TCN (A) & \textbf{59.6\%} & \textbf{71.1\%} & \textbf{50.7\%} & \textbf{61.3\%} & \textbf{54.3\%} & \textbf{65.0\%} & \textbf{38.5\%} & \textbf{49.2\%} \\ 

\bottomrule
\end{tabular}
\end{small}

\end{table}

\subsection{The Impact of Base Learner}
Table~\ref{tab:base_learner} shows the empirical comparison between bidirectional LSTM and TCN when ATAML is used as the meta learner.
The results suggest that TCN performs better than bidirectional LSTM across all experiments on miniReuters-21578.

\begin{table}[h]
\renewcommand{\arraystretch}{1.15}
\centering
\caption{Comparing bidirectional LSTM and TCN as base learners on miniReuters-21578}\smallskip
\centering
\label{tab:base_learner}
\begin{small}
\begin{tabular}{llcccccccc}
\toprule
\multicolumn{2}{c}{Method }                   & \multicolumn{2}{c}{5-way Micro-F1} & \multicolumn{2}{c}{10-way Micro-F1}  & \multicolumn{2}{c}{5-way Macro-F1} & \multicolumn{2}{c}{10-way Macro-F1}
\\\cmidrule(r{4pt}){1-2} \cmidrule(r{4pt}){3-4} \cmidrule(l){5-6} \cmidrule(r{4pt}){7-8} \cmidrule(l){9-10}
             Meta & Base           & 1-shot     & 5-shot     & 1-shot     & 5-shot  & 1-shot     & 5-shot     & 1-shot     & 5-shot    \\ \midrule
ATAML & LSTM (A) & 38.0\% & 62.3\% & 27.1\% & 33.7\% & 30.3\% & 50.2\% & 18.8\% & 21.2\% \\ 
ATAML & TCN (A) & \textbf{59.8\%} & \textbf{71.1\%} & \textbf{50.7\%} & \textbf{61.3\%} & \textbf{54.3\%} & \textbf{65.0\%} & \textbf{38.5\%} & \textbf{49.2\%} \\ 
\bottomrule
\end{tabular}
\end{small}

\end{table}
\subsection{Other Baseline Methods}
Table~\ref{tab:classic_baselines} shows the comparison between the proposed ATAML and classic machine learning methods, i.e., SVM, Naive Bayes Multinomial and KNN, which uses tfidf features as model inputs.
The results suggest that SVM and naive Bayes multinomial severely overfit on the training data generalizes poorly on evaluation.
The K-nearest neighbor classifier performs better than SVM and naive Bayes multinomial mainly because it is an nonparametric and distance-based algorithm.
The proposed ATAML is significantly better than KNN on the Micro-F1 measure and ATAML performs at least as good as KNN on the Macro-F1 measure.
\begin{table}[h]
\renewcommand{\arraystretch}{1.15}
\centering
\caption{Comparing ATAML with SVM, Naive Bayes Multinomial and KNN on miniReuters-21578}\smallskip
\centering
\label{tab:classic_baselines}
\begin{small}
\begin{tabular}{lcccccccc}
\toprule
\multicolumn{1}{c}{Method }                   & \multicolumn{2}{c}{5-way Micro-F1} & \multicolumn{2}{c}{10-way Micro-F1}  & \multicolumn{2}{c}{5-way Macro-F1} & \multicolumn{2}{c}{10-way Macro-F1} \\
 \cmidrule(r{4pt}){2-3} \cmidrule(l){4-5} \cmidrule(r{4pt}){6-7} \cmidrule(l){8-9}
                        & 1-shot     & 5-shot     & 1-shot     & 5-shot  & 1-shot     & 5-shot     & 1-shot     & 5-shot    \\ \midrule
SVM & 3.8\% & 35.8\% & 0.3\% & 18.8\%  & 3.3\% & 25.1\% & 0.2\% & 12.6\% \\ 
Naive Bayes Multinomial & 0.5\% & 7.7\% & 0.0\% & 0.0\%  & 0.2\% & 3.4\% & 0.0\% & 0.0\% \\ 
KNN & 46.7\% & 54.4\% & 39.4\% & 57.3\%  & 43.8\% & 37.3\% & \textbf{37.4\%} & \textbf{52.5\%} \\
ATAML, TCN (A) & \textbf{59.8\%} & \textbf{71.1\%} & \textbf{50.7\%} & \textbf{61.3\%} & \textbf{54.3\%} & \textbf{65.0\%} & \textbf{38.5\%} & \textbf{49.2\%} \\

\bottomrule
\end{tabular}
\end{small}
\end{table}

Table~\ref{tab:document_embedding} summarizes the comparison between the proposed ATAML and document embedding approaches, i.e., doc2vec~\cite{levine1985effect} and doc2vecC~\cite{chen2017efficient}.
In contrast to ATAML that uses attention to aggregate information from substructures of some text input, the document embedding approaches directly encode each document into one embedding vector and another classifier, such as KNN or SVM, is applied on the document embeddings for classification.

The empirical results suggest the document embedding approaches are not as effective as the proposed ATAML method.
This finding confirms the need to apply attention on substructures of text data, rather than treating each document as a static embedding vector.

\begin{table}[h]
\renewcommand{\arraystretch}{1.15}
\centering
\caption{Comparing ATAML with document embeddings methods on miniReuters-21578}\smallskip
\centering
\label{tab:document_embedding}
\begin{small}
\begin{tabular}{lcccccccc}
\toprule
\multicolumn{1}{c}{Method }                   & \multicolumn{2}{c}{5-way Micro-F1} & \multicolumn{2}{c}{10-way Micro-F1}  & \multicolumn{2}{c}{5-way Macro-F1} & \multicolumn{2}{c}{10-way Macro-F1} \\
 \cmidrule(r{4pt}){2-3} \cmidrule(l){4-5} \cmidrule(r{4pt}){6-7} \cmidrule(l){8-9}
                        & 1-shot     & 5-shot     & 1-shot     & 5-shot  & 1-shot     & 5-shot     & 1-shot     & 5-shot    \\ \midrule
Doc2Vec, KNN & 31.4\% & 42.0\% & 19.4\% & 32.9\% & 18.5\% & 28.9\% & 10.1\% & 22.5\% \\
Doc2Vec, SVM & 27.4\% & 59.1\% & 11.4\% & 44.3\% & 19.9\% & 44.6\% & 8.5\% & 31.0\% \\
Doc2VecC, KNN & 42.8\% & 62.6\% & 30.2\% & 50.0\% & 34.9\% & 53.2\% & 23.9\% & 42.2\% \\
Doc2VecC, SVM & 33.7\% & 58.4\% & 18.6\% & 42.7\% & 25.8\% & 46.0\% & 12.5\% & 30.3\% \\
ATAML, TCN (A) & \textbf{59.6\%} & \textbf{71.1\% }& \textbf{50.7\%} & \textbf{61.3\%} & \textbf{54.3\%} & \textbf{65.0\%} & \textbf{38.5\%} & \textbf{49.2\%} \\

\bottomrule
\end{tabular}
\end{small}

\end{table}

\section{Visualizing Attention}
\begin{figure}[!htbp]
  \centering
  \includegraphics[scale=0.68]{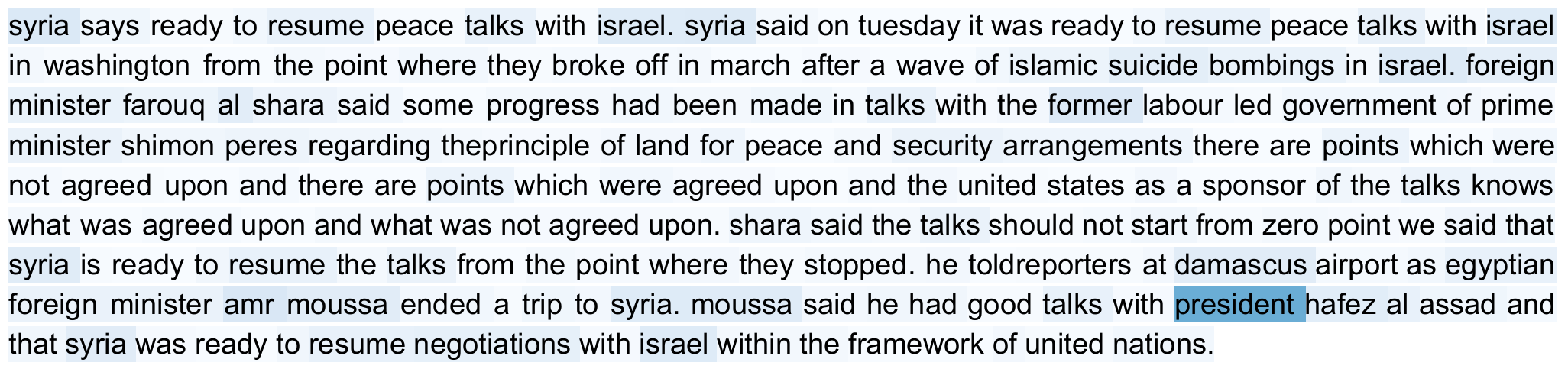}
  \caption{Visualizing attentions learned by MAML TCN(A).}
    \label{fig:attention_MAML_training}
\end{figure}

\begin{figure}[!htbp]
  \centering
  \includegraphics[scale=0.68]{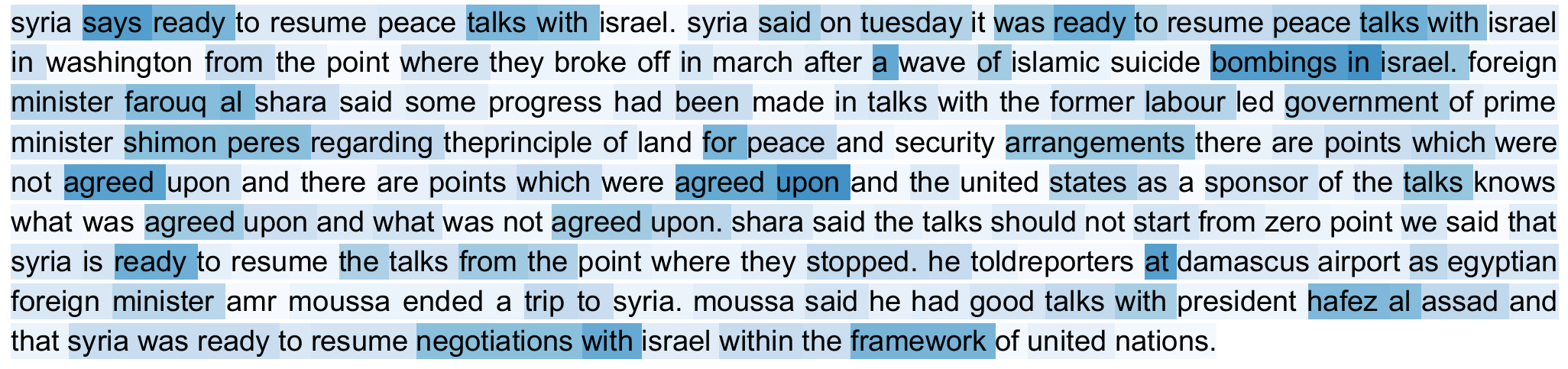}
  \caption{Visualizing attentions learned by ATAML TCN(A).}
    \label{fig:attention_ATAML_training}
\end{figure}

Figure~\ref{fig:attention_MAML_training} and Figure~\ref{fig:attention_ATAML_training} illustrate the the same training example after the meta-learned is trained with MAML and ATAML, respectively.
The target label for this document is ``INTERNATIONAL RELATIONS'' and both models make correct predictions for this training example.
Whereas the MAML model illustrated in Figure~\ref{fig:attention_MAML_training} is over-fitting to the keyword ``president'', the proposed ATAML model in Figure~\ref{fig:attention_ATAML_training} identifies multiple key phrases, such as ``talk with'', ``agrred upon'' and ``negotiation with'', that are important to the classification of ``INTERNATIONAL RELATIONS''.
Learning meaningful phrase-level representations regularizes a model from over-fitting to spurious correlation in the training examples.

\end{appendices}

\end{document}